\documentclass{article} 
\pdfoutput=1

\usepackage[utf8]{inputenc} 
\usepackage[T1]{fontenc}    
\usepackage{authblk}
\PassOptionsToPackage{hyphens}{url} 
\usepackage[hyperfootnotes=false]{hyperref}

\usepackage{enumitem}
\usepackage[sort&compress,numbers]{natbib}
\usepackage{mathtools} 
\usepackage{booktabs} 
\usepackage{microtype}
\usepackage{graphicx}
\usepackage{gincltex}
\usepackage{caption}
\usepackage{subcaption}
\usepackage{amsmath}
\usepackage{amssymb}
\usepackage{mathtools}
\usepackage{amsthm}
\usepackage{bbm}
\usepackage{dsfont}
\usepackage{amsthm}
\usepackage{nicematrix}
\usepackage{nicefrac}

\usepackage{stackengine}
\usepackage{fullpage}

\usepackage{multicol}
\usepackage{multirow}
\usepackage{makecell}
\usepackage{xcolor,colortbl}
\usepackage{tcolorbox}  
\usepackage{tikz}
\usetikzlibrary{patterns}

\newcommand{\stripedboxinline}[1]{%
\tikz[baseline=(eq.base)]{
    \node[inner sep=2pt] (eq) {$#1$};
    \draw[pattern=north west lines, pattern color=gray!20] (eq.south west) rectangle (eq.north east);
    \draw[gray] (eq.south west) rectangle (eq.north east);
    \node[inner sep=2pt] (inside) {$#1$};
        }%
}

\newcommand{\symbolBP}{\stripedboxinline{\color{boxgray}\textcolor{circlegray}{\bullet}\textcolor{redarrow}{\rightarrow} \textcolor{bluearrow}{\leftarrow}\textcolor{circlegray}{\bullet}}\,\textcolor{boxgray}{\scriptsize R}\,}

\newcommand{\symbolB}{$\color{boxgray}\textcolor{circlegray}{\bullet}\textcolor{redarrow}{\rightarrow} \textcolor{bluearrow}{\leftarrow}\textcolor{circlegray}{\bullet}\ $}

\newcommand{\symbolA}{$\color{boxgray}\textcolor{bluearrow}{\leftarrow} \textcolor{circlegray}{\bullet}\,\textcolor{circlegray}{\bullet}\textcolor{redarrow}{\rightarrow}\ $}

\newcommand{\symbolC}{$\color{boxgray} \textcolor{circlegray}{\bullet}\quad\textcolor{circlegray}{\bullet}\ $}

\newcommand{\given}{\,\mid\,}
\definecolor{Light}{rgb}{1.0, 0.95, 0.95}
\definecolor{Dark}{rgb}{1.0, 0.9, 0.9}
\definecolor{redarrow}{HTML}{c55c41}
\definecolor{bluearrow}{HTML}{4b869a}
\definecolor{boxgray}{HTML}{808082}
\definecolor{circlegray}{HTML}{606671}

\definecolor{A}{HTML}{bfedcd}
\definecolor{C}{HTML}{d8e5ff}
\definecolor{B}{HTML}{fcdcda}
\definecolor{BP}{HTML}{beeef0}

\usepackage{Definitions}

\title{Human-Alignment Influences \\ the Utility of AI-assisted Decision Making}

\author[1,2]{Nina~L.~Corvelo~Benz}
\author[1]{Manuel~Gomez~Rodriguez}
\affil[1]{Max Planck Institute for Software Systems, \{ninacobe, manuelgr\}@mpi-sws.org
}
\affil[2]{Department of Biosystems Science and Engineering, ETH Zurich
}

\date{}

\begin{document}

\maketitle


\begin{abstract}
Whenever an AI model is used to predict a relevant (binary) outcome in AI-assisted decision making, it is widely agreed that, together with each prediction, the model should provide an AI confidence value.
%
However, it has been unclear why decision makers have often difficulties to develop a good sense on when to trust a prediction using AI confidence values.
Very recently, Corvelo Benz and Gomez Rodriguez have argued that, for rational decision makers, the utility of AI-assisted decision making is inherently bounded by the degree of alignment between the AI confidence values and the decision maker's confidence on their own predictions.
%
In this work, we empirically investigate to what extent the degree of alignment actually influences the utility of AI-assisted decision making.
%
To this end, we design and run a large-scale human subject study ($n = 703$) where participants solve a simple decision making task---an online card game---assisted by an AI model with a steerable degree of alignment. 
%
Our results show a positive association between the degree of alignment and the utility of AI-assisted decision making. 
In addition, our results also show that post-processing the AI confidence values to achieve multicalibration with respect to the participants' confidence on their own predictions increases both the degree of alignment and the utility of AI-assisted decision making.
\end{abstract}


\section*{Introduction}
\label{sec:introduction}
%
State-of-the-art AI models have matched, or even surpassed, human experts at pre\-dicting relevant outcomes for decision making in a variety of high-stakes domains such as medicine, education and science~\cite{jiao2020deep,whitehill2017mooc,davies2021advancing}.
Consequently, the con\-ven\-tio\-nal wisdom is that human experts using these AI models should make \emph{better} decisions than those not using them.
However, multiple lines of empirical evidence suggest that this is not generally true~\cite{yin2019understanding,zhang2020effect,suresh2020misplaced,lai2023towards}.

In the canonical case of binary outcomes and binary decisions, Corvelo Benz and Gomez Rodriguez~\cite{corvelo2024human} have recently argued that the way in which AI models quan\-ti\-fy and communi\-cate confidence in their predictions may be one of the reasons AI-assisted decision making falls short.
Their key argument is that, if an AI model uses calibrated estimates of the probability that the predicted outcome matches the true outcome as AI confidence values, as commonly done~\cite{guo2017calibration, zadrozny2001obtaining, Gneiting2007, gupta2021distribution,huang2020tutorial,wang2022improving}, 
a rational human expert who places more (less) trust on predictions with higher (lower) AI confidence 
may never make optimal decisions.
%
%
Further, they show that the optimality gap is proportional to the maximum alignment error (MAE) between the AI confidence and the human expert'{}s confidence, \ie,
%
\begin{equation}\label{eq:alignment}
    \text{MAE} = \max_{h \leq h', a \leq a'} P(Y = 1 \given A=a, H=h) - P(Y = 1 \given A=a', H=h'),
\end{equation}
where $A$ and $H$ are the AI and the expert'{}s confidence values in the outcome $Y=1$, respectively.

While the above theoretical results are illuminating, they do not elu\-ci\-date to what extent there is an association between the degree of alignment between the AI confidence and the human expert'{}s confidence and the utility of AI-assisted decision making---they just show that the maximum utility we can expect from AI-assisted decision making is inherently bounded by the degree of alignment.
In this work, we aim to fill this gap by designing and running a large-scale human subject study where participants solve a simple decision making task---a card game---assisted by an AI model with a steerable degree of alignment.

\begin{figure}
    \centering
    \includegraphics[width=0.7\linewidth]{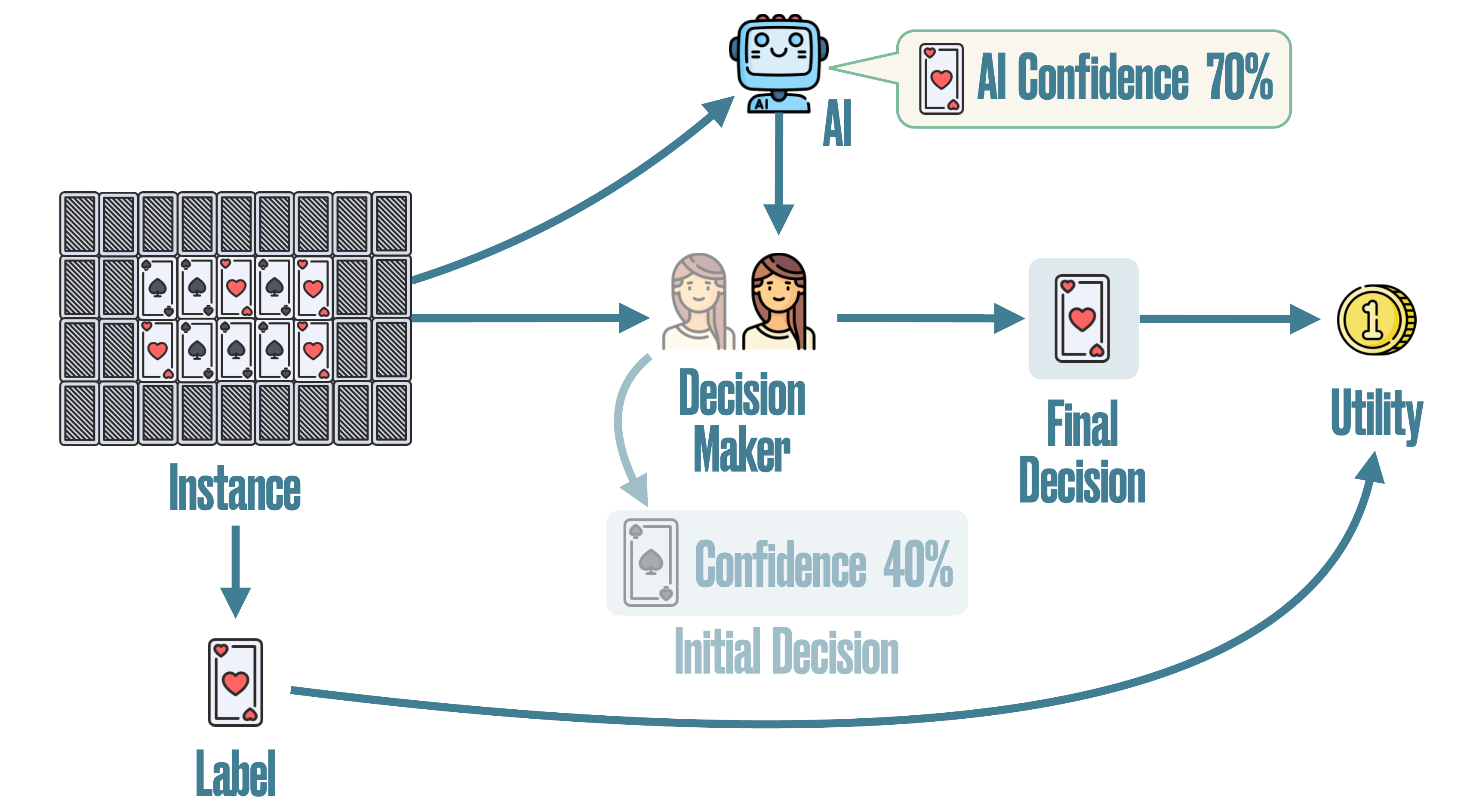}
    \caption{Our simple AI-assisted decision making task.}
    \label{fig:ai-assisted-dm}
\end{figure}

Our AI-assisted card game does not require prior knowledge besides a very basic understanding of probabilities, participants just need to guess the color of a ran\-dom\-ly picked card from a pile of red and black cards that is partially observed assisted by an AI model, as shown in Figure~\ref{fig:ai-assisted-dm}. 
To control the degree of alignment between the AI confidence and the participants' confidence, our game biases the proportion of red and black cards shown to the participants according to the AI confidence value.
%

Participants in our human subject study are randomly assigned to one of four different groups ($n\approx 100$ participants per group). 
In the first three groups, denoted by the symbols \symbolB, \symbolA and \symbolC, participants are assisted by the same AI model, which is perfectly calibrated; however, the degree of alignment varies across groups. 
In the fourth group, denoted by symbol \symbolBP, participants are assisted by an AI model whose predictions are post-processed to increase their degree of alignment via multicalibration using additional (held-out) data ($n=302$ participants) from the group with the lowest degree of alignment within the first three groups. 
We evaluate the impact of the degree of alignment on the utility of AI-assisted decision making by running multiple Bayesian A/B tests for two categories of games characterized by participants' initial performance and for pairs of groups \symbolB, \symbolA, \symbolC, and \symbolBP.
We find decisive evidence that the utility is greater in groups with higher degree of alignment compared to the group with the lowest degree of alignment for games where participants initially perform badly (Evidence Ratios $>100$ with an estimated difference in utility of greater or equal to $0.15$ between groups compared).
Moreover, we also find evidence that increasing the degree of alignment via multicalibration increases the utility achieved by participants in the fourth group  compared to the reference group before multicalibration for all games (Evidence Ratios of $2.59$ and $>100$ with an estimated difference in utility of $0.02$ between the two groups for each category of game).
%

%
%
%
%

\section*{AI-Assisted Game Design} 
\label{sec:game-design}
The game consists of different rounds $i\in \{1,\dots, 24\}$. 
In each round, the game shows the participant $21$ cards out of a game pile of $65$ cards and picks one card at random from the pile. 
Based on the cards shown, it asks the participant to guess the color of the card picked and give their (initial) confidence that the card picked is red. 
Then, it shows the participant the AI confidence that the card picked is red and allows them to revise their guess and confidence. 
Finally, the participant wins a point if their final guess is correct and otherwise they lose a point. 
To incentivize participants, we give a bonus compensation of $\pounds 0.12$ at the end of the game for each point earned.

\subsection*{A perfectly calibrated AI, by design}
Participants in the first three groups are shown $24$ different types of game piles $S$, a different type per round in randomized order. Each type of game pile $(r, a) \in S$ is de\-fined by the fraction $r$ of red cards in the game pile and the AI confidence $a$.
The AI confidence $a$ takes values from the set $\Acal = \{1/13, 2/13 \dots, 12/13\}$ and the fraction $r$ of red cards is given by
\[r \in \{ a-\text{var}_a, a+\text{var}_a\},\]
where $\text{var}_a$ is chosen such that, in Eq.~\ref{eq:wallenius},  $r\cdot 65$ is an integral number of cards for all $r$.
To ease interpretability, participants are shown re-scaled AI confidence values rounded to the nearest integer between $0$ and $100$.
As $r$ determines the probability that the game picks a red card from the game pile and each participant is shown each type of game pile once,
%
the AI confidence is perfectly calibrated, \ie, for all $a \in \Acal$, it holds that
\begin{equation*}
    P(C=\texttt{red} \given A=a) = a\,. 
\end{equation*}
We also validated empirically that the rounded AI confidence values indeed obtain a low calibration error as well (see Appendix Table~\ref{tab:calibration-error}).

Further, 
since one can argue that the AI model has essentially 
information about only $13$ cards whereas participants have information about $21$ cards, the latter have more information about each pile than the former.

\subsection*{Steering alignment by biasing the proportion of red and black cards shown to the participants}


%
In each round, the $21$ cards of the game pile shown to the participants in the first three groups are sampled from a Wallenius' noncentral hypergeometric distribution with an odds ratio $\omega_g(r ,a )$ defined by the group $g$ participants are assigned to. 

\begin{figure}[t!]
    \centering
    \includegraphics[width=\linewidth]{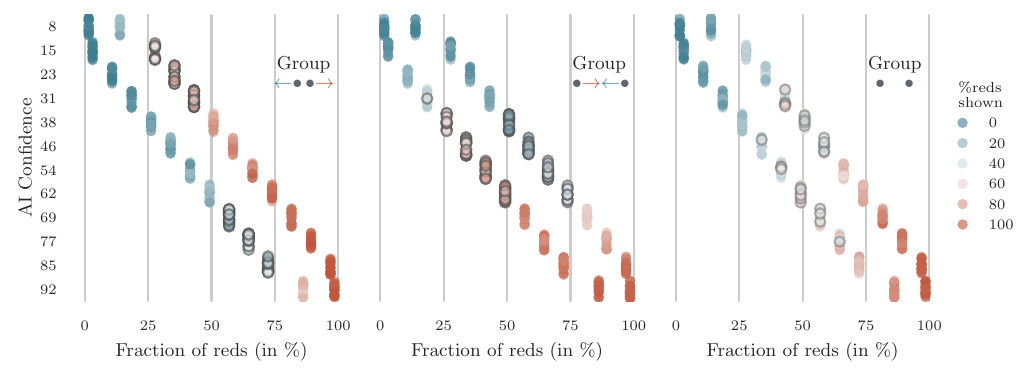}
\caption{
AI confidence $a$ and fraction of red cards shown $z$ against the fraction of red cards $r$ per game played by participants in each group. Games in which the participant are most likely mislead since the majority of cards shown have the opposite of the overall majority color are highlighted in gray. 
}\label{fig:generated-games}
\end{figure}

The odds ratio $\omega_g(r ,a )$ biases the likelihood of the color ratio in the cards shown to the participants---if $\omega_g(r ,a )$ is greater (smaller) than $1$, it is more likely to sample a red (black) card than a black (red).
More formally, the fraction $z $ of red cards shown to the participant is given by
\begin{equation} \label{eq:wallenius}
    z  \sim \text{wnchypg}\left(21, 65\cdot r , 65\cdot(1-r ), \omega_g(r ,a )\right).
\end{equation}
and, for each group $g$, the odds ratio $\omega_g(r ,a )$ biases the generation as follows:
\begin{itemize}
    \item \textbf{Group \symbolB}: Bias the fraction of reds shown towards the AI confidence---the odds of sampling reds (blacks) is greater than the odds of sampling blacks (reds) when the true probability $r $ is lower (greater) than the AI confidence $a $.
    We set \[\omega_g(r ,a )= \begin{cases}
        1/4 \quad \text{if } r  > a  , \\
        4 \quad \text{if } r  < a   .
    \end{cases}\]
    \item \textbf{Group \symbolA}: Bias the fraction of reds shown away from the AI confidence---the odds of sampling reds (blacks) is greater than the odds of sampling blacks (reds) when the true probability $r $ is greater (lower) than the AI confidence $a $.
    We set \[\omega_g(r ,a )= \begin{cases}
        4 \quad \text{if } r  > a  , \\
        1/4 \quad \text{if } r  < a  .
    \end{cases}\]
    \item \textbf{Group \symbolC}: No bias.
    We set $\omega_g(r ,a )= 1 $.
\end{itemize}
%
The resulting empirical distribution of red cards shown to the participants across game piles is shown in Figure~\ref{fig:generated-games}.
In Group \symbolB, we bias the fraction of reds shown towards the AI confidence, when conditioning on the AI confidence. 
Therefore, we expect that participants have lower (higher) confidence in guessing red for games where the true probability of picking a red card in the game pile is higher (lower) than the AI confidence leading to lower degree of alignment (see Eq.~\ref{eq:alignment}). 
However, in Group \symbolA and \symbolC, we expect the opposite to happen and, thus, the AI to have a higher degree of alignment with  participants.
Here, note that, the quantity and type of games where participants may perform poorly are different across groups. In particular, the unbiased Group \symbolC has fewer game instances compared to the groups \symbolA and \symbolB in which the participant may be misled by the color ratio of the cards shown.

\subsection*{Increasing alignment via multicalibration}

Participants in the Group \symbolBP are shown the same type of game piles with the same fraction $z$ of red cards shown as participants 
in the Group~\symbolB.
However, the AI confidence is post-processed by the uniform mass histogram binning algorithm introduced by Corvelo Benz and Gomez Rodriguez~\cite{corvelo2024human} for the purpose of multicalibration. 
We run the algorithm with additional (held-out) calibration data from the Group \symbolB for a different set of games from the same distribution.
Refer to the Materials and Methods section and the Appendix Figure~\ref{fig:calibration-data} for more information regarding the post-processing algorithm and the calibration data.


\section*{Results} 
\label{sec:results}

\begin{figure*}[h!]
    \centering
    \includegraphics[width=1\linewidth]{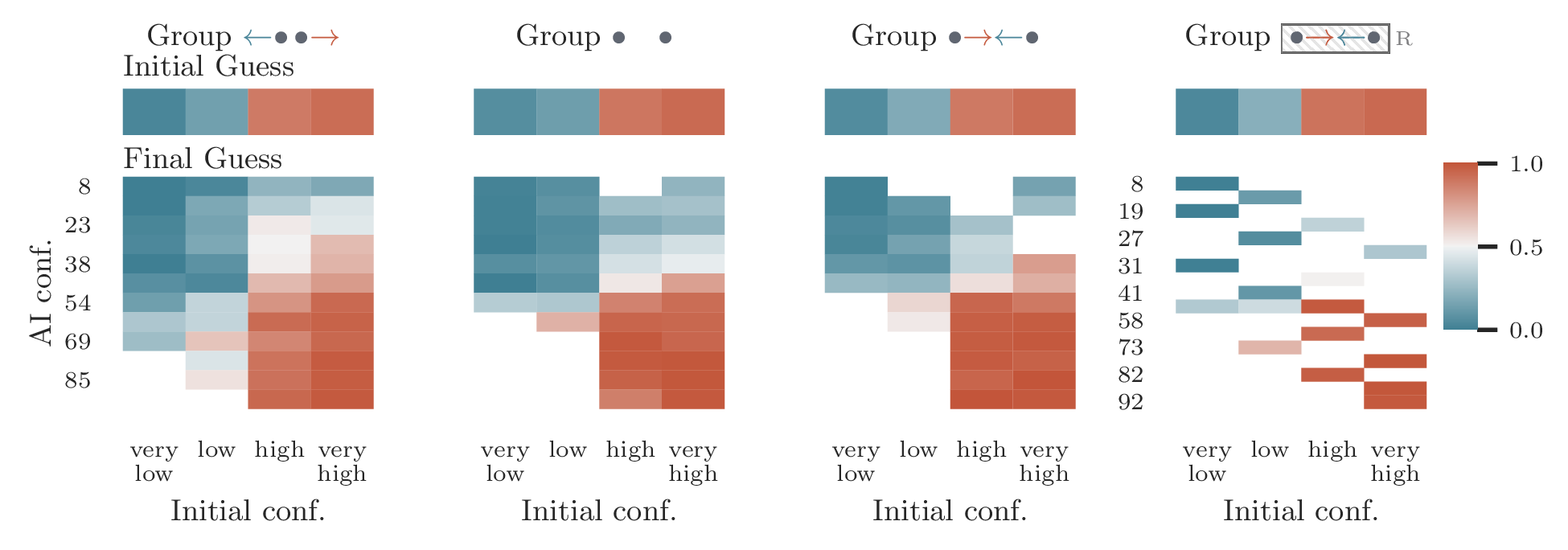}
    \caption{Heatmap of decision probabilities of guessing \texttt{red} stratified by participants' initial confidence and AI confidence shown, and averaged over participants. The initial confidence recorded is discretized into four bins---very low, low, high, very high---denoting the confidence of the participants that the color of the picked card will be red. Bins with 10 or less data points are not displayed. }
    \label{fig:prob-map}
\end{figure*}
We first quantify the degree of alignment between the AI conﬁdence and the human expert’s conﬁdence for each group by measuring the empirical maximum alignment error (MAE, Eq.~\ref{eq:alignment}) and expected alignment error (EAE, Eq.~\ref{eq:eae}).
Table~\ref{fig:alignment-error} summarizes the results.
We observe the lowest alignment error in Group \symbolC, closely followed by Group \symbolA, and the highest alignment error in Group \symbolB, where we observe more than twice higher MAE and a $20$ times higher EAE than in Group \symbolC. Further, in Group \symbolBP, we observe that the multicalibration algorithm helps to reduce the EAE by a third and the MAE by almost half in comparison to Group \symbolB.
In addition, in Figure~\ref{fig:histogram} in the Appendix, we also observe that, within each group, the initial confidence of participants is normally distributed around the percentage $z$ of reds shown. 
Based on these findings, we conclude that, as expected, our AI-assisted game design has successfully steered the degree of alignment in each group. 

Further, we verify that, on average, participants behave rationally in all groups, which is one of the motivating assumptions underpinning our work.
More specifically, in Figure~\ref{fig:prob-map}, we observe that the average probability that a participant'{}s initial guess is red depends monotonically on the participant'{}s own confidence that the card chosen 
is red and the probability that a participant'{}s final guess is red depends monotonically on both the participant'{}s own confidence and the AI confidence that the card chosen is red.  
As an immediate consequence, the maximum utility of AI-assisted decision making we can hope for is bounded by the degree of alignment in each group, although with minimal differences (see Appendix under Additional Experimental Results). 
%
\begin{table}[t!]
    \centering
    \caption{Expected Alignment Error (EAE) and Maximum Alignment Error (MAE) for each group.}
    \begin{NiceTabular}{cll}
    \CodeBefore
    \Body
           \bf \small Group & \bf \small EAE & \bf \small MAE  \\
     \hline
     \symbolA & 0.00065 & 0.1\\
     \symbolC & 0.0003 & 0.06\\
     \symbolB  & 0.00693 & 0.2\\
      \ \,\,\symbolBP & 0.00236 & 0.12\\
    \hline
    \end{NiceTabular} 
    \label{fig:alignment-error}
\end{table}

Next, we contrast the utility achieved by participants across groups 
controlling for the different quantities and types of game piles 
across groups. 
To this end, we compare the initial guess $d$ and the final guess $d'$ made by each participant in each game 
against the optimal guess $\pi^* = \pi^*(r)$, where $\pi^*(r )$ is red if and only if the fraction of red cards $r >0.5$ (there are no games with $r =0.5$). 
%
%
More formally, we focus on the conditional matching rates 
\[ \theta_1 = \EE[Q' \given Q =1] \quad \text{and} \quad \theta_0 = \EE[Q'  \given Q =0] \]
where the expectation is over games and participants and
    \[ Q  := \mathds{1}[\pi^*(r ) = d ] \quad \text{and} \quad 
    Q' := \mathds{1}[\pi^*(r ) = d' ].\]
The conditional matching rates measure how close participants are on average to the optimal decision for each game after observing the AI confidence given that they were initially optimal or not optimal.
Comparing conditional matching rates instead of other utility measures has two main benefits:
For one, this measure is more comparable between groups, because, for a given type of game pile, the optimal guess is the same across groups for each participant since $\pi^*$ depends only on $r$. Furthermore, because we compare rates conditional on the initial performance $Q$ of the participants, we take into account the effect of different amounts of games in each group where participants are misled.

To evaluate the impact of the degree of alignment on the participants' conditional matching rates in each group, we conduct multiple Bayesian A/B tests: We first test if the conditional matching rates $\theta_0$ and $\theta_1$ are greater for groups \symbolA and \symbolC with higher degree of alignment than for Group \symbolB with lower degree of alignment. We then test if the conditional matching rates $\theta_0$ and $\theta_1$ are greater after post-processing for Group \symbolBP than for Group \symbolB. To this end, we fit two Bayesian binomial (logit) mixed effects models for the (unconditional) matching rate $\theta = \EE[Q']$ with an interaction coefficient between the group condition $g$ and $Q$, and a random intercept for each participant, \ie, 
\[
\log\left(\frac{\theta}{1-\theta}\right) \sim 0 + g * Q + (1 \given \text{participant}).\] 
We opt for no fixed global intercept in order to not have a fixed reference group in the model. However, fitting the model with a fixed global intercept returns equivalent results. Note that, the conditional matching rates $\theta_0$ and $\theta_1$ can be recovered from the model by setting $Q=0$ and $Q=1$, respectively.

\begin{figure}
%
        \begin{minipage}{0.5\textwidth}
        \centering
        \includegraphics[width=\linewidth]{./img/bayes_ABC_initialmatch_0.tex}
        \\
        \subfloat[Results of Bayes A/B test when initial match $Q =0$ for groups \symbolA, \symbolC and \symbolB: The plot shows the Bayes posterior of  conditional matching rate $\theta_0$ for all three groups. The table shows the estimated difference in $\theta_0$, the estimation error, the evidence ratio (Bayes factor) and the posterior probability for the hypothesis stating that $\theta_0$ is larger in Group \symbolA (or Group \symbolC) than in Group \symbolB.]{
            \resizebox{0.9\linewidth}{!}{%
            \begin{NiceTabular}{ccccc}
            \CodeBefore
            \rowcolor{Light}{1}
            \rowcolor{Dark}{2}
            \rowcolor{Light}{3}
            \columncolor{white}{1}
            \Body
            \centering
                  \bf \small Groups & \bf \small Est. & \bf \small \makecell{Est. \\Error} & \bf \small \makecell{Evid. \\Ratio}  & 
              \bf \small \makecell{Post.\\Prob.} \\
            \symbolA\, vs. \symbolB  &  0.16 & 0.03 & $\infty$ & 1.00  \\
            \symbolC\, vs. \symbolB &   0.15 & 0.03  & $\infty$ & 1.00\\
        \end{NiceTabular}
        }
        \label{fig:bayes-results-ABC-0}
        }
        \end{minipage}
        \begin{minipage}{0.5\textwidth}
        \centering
        \includegraphics[width=\linewidth]{./img/bayes_ABC_initialmatch_1.tex}
        \\
        \subfloat[Results of Bayes A/B test when initial match $Q =1$ for groups \symbolA, \symbolC and \symbolB: The plot shows the Bayes posterior of  conditional matching rate $\theta_1$ for all three groups. The table shows the estimated difference in $\theta_1$, the estimation error, the evidence ratio (Bayes factor) and the posterior probability for the hypothesis stating that $\theta_1$ is larger in Group \symbolA (or Group \symbolC) than in Group \symbolB.]{
            \resizebox{0.9\linewidth}{!}{%
            \begin{NiceTabular}{ccccc}
            \CodeBefore
            \rowcolor{Light}{1}
            \rowcolor{Dark}{2}
            \rowcolor{Light}{3}
            \columncolor{white}{1}
            \Body
                  \bf \small  Groups & \bf \small Est. & \bf \small \makecell{Est. \\Error} & \bf \small \makecell{Evid. \\Ratio} & \bf \small \makecell{Post.\\Prob.} \\
            \symbolA\, vs. \symbolB & -0.01 & 0.01 & 0.05  & 0.05 \\
            \symbolC\, vs. \symbolB & -0.02 & 0.01 & 0.01  & 0.01 \\
        \end{NiceTabular}
        }
        \label{fig:bayes-results-ABC-1}
        }
        \end{minipage}
        \\
        \subfloat[Empirical matching rate of participants with the optimal guess stratified by initial matching rate and whether the initial guess agrees with the AI confidence. Bins with 10 or less data points are omitted for sake of clarity.]{
        \centering
        \includegraphics[width=0.9\linewidth]{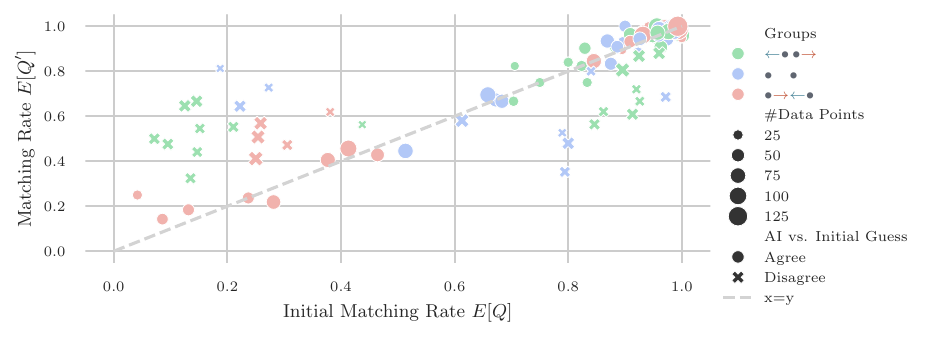}
        \vspace{-1em}
        \label{fig:consensus-ABC}
        }
        \caption{Results for groups \symbolA, \symbolC and \symbolB.}
    \end{figure}
Figures~\ref{fig:bayes-results-ABC-0} and~\ref{fig:bayes-results-ABC-1} show the results of the Bayesian A/B tests for the groups \symbolA, \symbolC and \symbolB. We observe that the support of the Bayes posterior for $\theta_0$ in groups \symbolA and \symbolC obtains higher values than in Group \symbolB. Both A/B tests also show decisive evidence to support the hypothesis that $\theta_0$ in Group \symbolA and in Group \symbolC is greater than in Group \symbolB (Evidence Ratio $>100$ for both tests), with the estimated difference in $\theta_0$ greater or equal to $0.15$ in both tests. The tests do not support the hypothesis that $\theta_1$ in groups \symbolA or \symbolC is greater than in Group \symbolB (Evidence Ratio $<1$ for both tests). However, the estimated difference in $\theta_1$ was only $-0.01$ for Group \symbolA and $-0.02$ for Group \symbolC. 
These findings suggest that, in groups with higher degree of alignment, the AI was more helpful in correcting the non-optimal guess made initially by the participants, while being slightly more harmful and misleading the participants when participants' initial guess was optimal.

For a more fine-grained understanding of how the degree of alignment influences the matching rates,
we stratified the empirical matching rate $\theta=\EE[Q']$ by the initial matching rate $\EE[Q]$ of the participants and whether the initial guess agrees with the AI confidence. The results are shown in
Figure~\ref{fig:consensus-ABC}. We observe that the regions of disagreement between AI guess and initial guess, where the AI helped or misled the participants, are larger in quantity and size for the groups with higher degree of alignment than for Group \symbolB. This suggests that the potential to learn a better guess is higher when the AI is more aligned with the participants. Although in Group \symbolA and \symbolC participants were misled by the AI for some games, we hypothesize that, if they would play more games, participants could eventually learn to trust their own confidence more for these games and rely on the AI confidence more for the games where they perform worse.
    \begin{figure}
    \begin{minipage}{0.5\textwidth}
            \centering
            \includegraphics[width=\linewidth]{./img/bayes_BBP_initialmatch_0.tex}
        \\
        \subfloat[Results of Bayes A/B test when initial match $Q =0$ for groups \symbolBP and \symbolB: The plot shows the Bayes posterior of conditional matching rate $\theta_0$ for both groups. The table shows the estimated difference in $\theta_0$, the estimation error, the evidence ratio (Bayes factor) and the posterior probability for the hypothesis stating that $\theta_0$ is larger in Group \symbolBP than in Group \symbolB.]{
            \resizebox{0.9\linewidth}{!}{%
            \begin{NiceTabular}{ccccc}
            \CodeBefore
            \rowcolor{Light}{1}
            \rowcolor{Dark}{2}
            \rowcolor{Light}{3}
            \columncolor{white}{1}
            \Body
                  \bf Groups & \bf \small Est. & \bf \small \makecell{Est. \\Error} & \bf \small \makecell{Evid. \\Ratio}  & \bf \small \makecell{Post.\\Prob.} \\
            \symbolBP \,vs. \symbolB  & 0.02 & 0.03 & 2.59  & 0.72 \\
        \end{NiceTabular}
        }
        \label{fig:bayes-results-BBP-0}
        }
        \end{minipage}
        \begin{minipage}{0.5\textwidth}
        \centering
        \includegraphics[width=\linewidth]{./img/bayes_BBP_initialmatch_1.tex}
        \\
        \subfloat[Results of Bayes A/B test when initial match $Q =1$ for groups \symbolBP and \symbolB: The plot shows the Bayes posterior of conditional matching rate $\theta_1$ for both groups. The table shows the estimated difference in $\theta_1$, the estimation error, the evidence ratio (Bayes factor) and the posterior probability for the hypothesis stating that $\theta_1$ is larger in Group \symbolBP than in Group \symbolB.]{
            \resizebox{0.9\linewidth}{!}{%
            \begin{NiceTabular}{ccccc}
            \CodeBefore
            \rowcolor{Light}{1}
            \rowcolor{Dark}{2}
            \rowcolor{Light}{3}
            \columncolor{white}{1}
            \Body
                  \bf Groups & \bf \small Est. & \bf \small  \makecell{Est. \\Error} & \bf \small \makecell{Evid. \\Ratio}  & 
              \bf \small \makecell{Post.\\Prob.} \\
            \symbolBP \,vs. \symbolB  & 0.02 & 0.01 & 113.94  & 0.99 \\
        \end{NiceTabular}
        }
        \label{fig:bayes-results-BBP-1}
        }
        \end{minipage}
        \\
        \subfloat[Empirical matching rate of participants with the optimal guess stratified by initial matching rate and whether the initial guess agrees with the AI confidence. Bins with 10 or less data points are omitted for sake of clarity.]{
        \centering
        \includegraphics[width=0.9\linewidth]{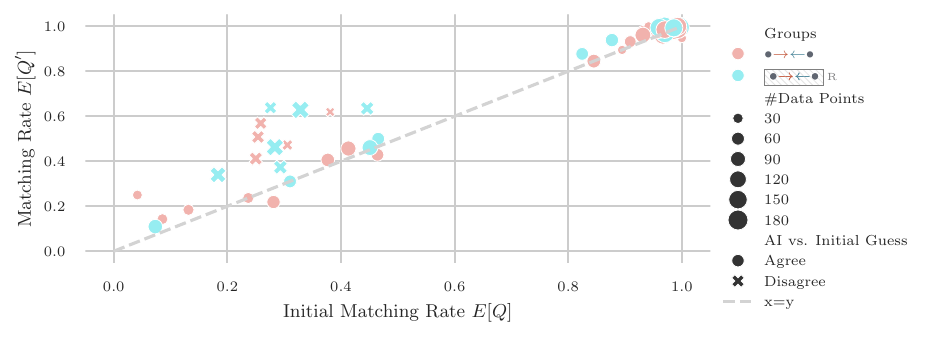}
        \label{fig:consensus-BBP}
        }
        \caption{Results for groups \protect\symbolBP and \symbolB.}
\end{figure}

Figures~\ref{fig:bayes-results-BBP-0} and~\ref{fig:bayes-results-BBP-1} show the results of the Bayesian A/B tests for the groups \symbolB and \symbolBP. We observe that the support of the Bayes posterior for $\theta_0$ and $\theta_1$ in Group \symbolBP obtains higher values than in Group \symbolB. While the A/B test only shows weak evidence to support the hypothesis that $\theta_0$ in Group \symbolBP is greater than in Group \symbolB (Evidence Ratio $2.59$), it shows decisive evidence to support the hypothesis that $\theta_1$ in Group \symbolBP is greater than in Group \symbolB (Evidence Ratio $>100$). However, the estimated difference in both $\theta_0$ and $\theta_1$ was $0.02$ showing only moderate improvement over Group \symbolB. Nevertheless, these findings suggest that the realigned AI was both more helpful in correcting the non-optimal guess made by the participants and less harmful when participants' initial guess was optimal. We hypothesize that this effect will be larger if the realignment of the AI decreases the alignment error further, for example, by using more fine grained bins during multicalibration or having more calibration data.

Figure~\ref{fig:consensus-BBP} shows the empirical matching rate $\theta=\EE[Q']$ stratified by the initial matching rate $\EE[Q]$ of the participants and whether the initial guess agrees with the AI confidence. We also observe for Group \symbolBP that the regions of disagreement between AI guess and initial guess are slightly more frequent than for Group \symbolB. This suggests that the potential for participants to learn a better guess is greater with the realigned AI. In addition, we also observe that the multicalibration process does not create regions where the participants are misled by the AI when they were initially optimal unlike the regions present in groups \symbolA and \symbolC.

\section*{Materials and Methods} 
\label{sec:materials}
Our data and analysis code are available at \url{https://github.com/Networks-Learning/human-alignment-study}. 
The study setup is available to play at
\url{https://hac-experiment.mpi-sws.org/?PROLIFIC_PID=test&STUDY_ID=test&SESSION_ID=test&LEVEL=B&GAME_BATCH=0}. 
The study reported in this article was approved by the institutional review board of ETH Zurich under Institutional Review Board Protocol EK 2024-N-49 ("User Study on Human-Aligned Calibration for AI-Assisted Decision Making"). All participants gave informed consent in advance.

\xhdr{Recruitment of Participants}
We recruited in total 703 participants on Prolific for our study (average age, 38.44 years; age range, 18 to 79 years; gender, 345 females, 8 non-binary, 4 not disclosed). Out of these participants, 302 participants were recruited to obtain calibration data for the multicalibration algorithm~\cite{corvelo2024human}. For the different groups, balanced condition assignment and repeat-participant exclusion were performed directly on Prolific. Upon joining the study, each participant was assigned to Group \symbolA (100 participants), Group \symbolC (99 participants), Group \symbolB (102 participants), or Group \symbolBP (100 participants).
The study included an instruction block, a practice game, three comprehesion/attention checks, 24 game instances, and a set of end of game and end of study surveys, as well as a demographic survey. More information about the surveys, as well as an overview of the study timeline, is provided in Figures~\ref{fig:flowchart} to~\ref{fig:demographic} in the Appendix.
The three attention checks were designed as simple game instances with a fraction of \texttt{reds} of $0\%,\ 75\%,$ and $100\%$ with no bias in the cards shown to the participants. The same three attention checks were used in each group, with the same configuration of cards shown to participants, and in the same order and place in the study timeline. We used the attention checks to filter out participants who performed out of the norm compared to most other participants: In more than one attention check, the confidence of the filtered participant was more than one standard deviation away from the mean confidence computed for each attention test over all groups. This resulted in on average 15.25 participants being excluded per group (15, 20, 17, and 9, respectively).
We perform the same filtering procedure with participants recruited for obtaining calibration data (the mean and standard deviation being computed over only these participants), resulting in the exclusion of 42 participants.

\xhdr{Game Batches}
For each group, we construct $20$ game batches with one game pile $(r,a,z)$ per type of game $(r,a)\in S$ by sampling $z$ from the Wallenius' non-central geometric distribution defined in Eq.~\ref{eq:wallenius}.
Similarly, we construct $60$ game batches using the same distribution as in Group \symbolB to gather calibration data.
Each participant completes one of the game batches consisting of $24$ games such that, for each game batch, we obtain data from $4$ to $6$ participants.

\xhdr{Metrics and Evaluation}
In each game, participants are asked to state their initial confidence that the card picked is \texttt{red} in a range from $0$ and $100$. We transform this recorded confidence into a discretized confidence $h \in \Hcal=\{\text{very low, low, high, very high}\}$ by dividing the confidence range into four regions of equal size ($[0,25]$, $[26,50]$, $[50,75]$, $[76,100]$). When the initial confidence is $50$, we assign the confidence value $\text{high}$ ($\text{low}$) if the initial guess is red (black).

To quantify the level of alignment error, we use the Maximum Alignment Error (MAE) and the Expected Alignment Error (EAE). The MAE is defined by Eq.~\ref{eq:alignment} and the EAE is defined as 
\begin{align}
\begin{split}\label{eq:eae}
\text{EAE} = \frac{1}{N} \cdot \sum_{h \leq h', a \leq a'} [ P(Y = 1 \given A=a, H=h) \\
 - P(Y = 1 \given A=a', H=h') ]_{+}, 
\end{split} 
\end{align}
where $N=|\{h \leq h', a \leq a'\}|$.

To run the Bayesian A/B tests, we use the "brms" package in R to fit the following model type
\begin{equation*}
    Q' \given \text{trials}(Q') \sim 0 + g * Q + (1 \given \text{participant})
\end{equation*}
using the binomial family. By default, this fits a logistic model for the proportions $\frac{\theta}{1-\theta}$. We set an uninformative standard normal prior $\Ncal(0,1)$ for all parameters of the model. We fit two different models: one for groups \symbolA, \symbolC, and \symbolB, and one for groups \symbolB and \symbolBP.
The Bayesian leave-one-out cross-validation estimate of the expected log pointwise predictive probabilities (ELPD) for the former model is $-815.2$ (standard error $26.6$, $\#$observations $497$) and for the latter model is $-547.7$ (standard error $23.4$, $\#$observations $352$). For comparison, a null model with uniform distribution across 25 possible outcomes (0-24) would obtain an ELPD of $\log(1/25)=-3.22$ per observation, so our models---with ELPD of $-1.64$ and $-1.55$ per observation---perform better than uniform random guessing.
Given the fitted models, we use the \texttt{hypothesis} function of "brms" to run the following Bayesian A/B tests: $\theta_{0, \text{\symbolA}}> \theta_{0, \text{\symbolB}}$, $\theta_{1, \text{\symbolA}}> \theta_{1, \text{\symbolB}}$, $\theta_{0, \text{\symbolC}}> \theta_{0, \text{\symbolB}}$, $\theta_{1, \text{\symbolC}}> \theta_{1, \text{\symbolB}}$, $\theta_{0, \text{\symbolBP}}> \theta_{0, \text{\symbolB}}$ and $\theta_{1, \text{\symbolBP}}> \theta_{1, \text{\symbolB}}$ where the second subscript indicates the group condition.

\xhdr{Realignment Algorithm}
Since participants' confidence partitions the game instances into disjoint subspaces, we  realign the AI model using the multicalibration algorithm based on uniform mass binning outlined by Corvelo Benz and Gomez Rodriguez~\cite{corvelo2024human}.
The algorithm partitions each subspace into $N=5$ uniform mass bins based on the AI confidence. Then, it computes the empirical mean fraction of reds for those game instances in the same bin. This empirical mean is computed using the true fraction of reds of each game instance as it gives more reliable estimates than using Bernoulli samples for small sample size.
Finally, the algorithm returns this empirical estimate as the re-aligned AI confidence for new game instances in the same subspace---with the same participant's confidence---falling into the same AI confidence bin.  
Since the AI confidence obtains discrete levels in our study, we add a small random uniform noise (in $[0,\nicefrac{1e-10}{1+1e-10}]$) to the AI confidence, both during training and deployment for Group \symbolBP, for the theoretical guarantees of the algorithm to hold.

\vspace{2mm}
\xhdr{Acknowledgements} Gomez-Rodriguez acknowledges support from the European Research Council (ERC) under the European Union'{}s Horizon 2020 research and innovation programme (grant agreement No. 945719).

{ 
\small
\bibliographystyle{unsrtnat}
\bibliography{human-alignment-study}

\begin{thebibliography}{14}
\providecommand{\natexlab}[1]{#1}
\providecommand{\url}[1]{\texttt{#1}}
\expandafter\ifx\csname urlstyle\endcsname\relax
  \providecommand{\doi}[1]{doi: #1}\else
  \providecommand{\doi}{doi: \begingroup \urlstyle{rm}\Url}\fi

\bibitem[Jiao et~al.(2020)Jiao, Atwal, Polak, Karlic, Cuppen, Danyi, de~Ridder, van Herpen, Lolkema, Steeghs, et~al.]{jiao2020deep}
Wei Jiao, Gurnit Atwal, Paz Polak, Rosa Karlic, Edwin Cuppen, Alexandra Danyi, Jeroen de~Ridder, Carla van Herpen, Martijn~P Lolkema, Neeltje Steeghs, et~al.
\newblock A deep learning system accurately classifies primary and metastatic cancers using passenger mutation patterns.
\newblock \emph{Nature communications}, 11\penalty0 (1):\penalty0 1--12, 2020.

\bibitem[Whitehill et~al.(2017)Whitehill, Mohan, Seaton, Rosen, and Tingley]{whitehill2017mooc}
Jacob Whitehill, Kiran Mohan, Daniel Seaton, Yigal Rosen, and Dustin Tingley.
\newblock Mooc dropout prediction: How to measure accuracy?
\newblock In \emph{Proceedings of the fourth (2017) acm conference on learning@ scale}, pages 161--164, 2017.

\bibitem[Davies et~al.(2021)Davies, Veli{\v{c}}kovi{\'c}, Buesing, Blackwell, Zheng, Toma{\v{s}}ev, Tanburn, Battaglia, Blundell, Juh{\'a}sz, et~al.]{davies2021advancing}
Alex Davies, Petar Veli{\v{c}}kovi{\'c}, Lars Buesing, Sam Blackwell, Daniel Zheng, Nenad Toma{\v{s}}ev, Richard Tanburn, Peter Battaglia, Charles Blundell, Andr{\'a}s Juh{\'a}sz, et~al.
\newblock Advancing mathematics by guiding human intuition with ai.
\newblock \emph{Nature}, 600\penalty0 (7887):\penalty0 70--74, 2021.

\bibitem[Yin et~al.(2019)Yin, Wortman~Vaughan, and Wallach]{yin2019understanding}
Ming Yin, Jennifer Wortman~Vaughan, and Hanna Wallach.
\newblock Understanding the effect of accuracy on trust in machine learning models.
\newblock In \emph{Proceedings of the {CHI} conference on human factors in computing systems}, pages 1--12, 2019.

\bibitem[Zhang et~al.(2020)Zhang, Liao, and Bellamy]{zhang2020effect}
Yunfeng Zhang, Q~Vera Liao, and Rachel~KE Bellamy.
\newblock Effect of confidence and explanation on accuracy and trust calibration in ai-assisted decision making.
\newblock In \emph{Proceedings of the Conference on Fairness, Accountability, and Transparency}, pages 295--305. ACM, 2020.

\bibitem[Suresh et~al.(2020)Suresh, Lao, and Liccardi]{suresh2020misplaced}
Harini Suresh, Natalie Lao, and Ilaria Liccardi.
\newblock Misplaced trust: Measuring the interference of machine learning in human decision-making.
\newblock In \emph{Proceedings of the ACM Conference on Web Science}, pages 315--324. ACM, 2020.

\bibitem[Lai et~al.(2023)Lai, Chen, Liao, Smith-Renner, and Tan]{lai2023towards}
Vivian Lai, Chacha Chen, Q~Vera Liao, Alison Smith-Renner, and Chenhao Tan.
\newblock Towards a science of human-ai decision making: An overview of design space in empirical human-subject studies.
\newblock In \emph{Proceedings of the 2023 ACM Conference on Fairness, Accountability, and Transparency}, 2023.

\bibitem[Corvelo~Benz and Rodriguez(2024)]{corvelo2024human}
Nina Corvelo~Benz and Manuel Rodriguez.
\newblock Human-aligned calibration for ai-assisted decision making.
\newblock \emph{Advances in Neural Information Processing Systems}, 36, 2024.

\bibitem[Guo et~al.(2017)Guo, Pleiss, Sun, and Weinberger]{guo2017calibration}
Chuan Guo, Geoff Pleiss, Yu~Sun, and Kilian~Q Weinberger.
\newblock On calibration of modern neural networks.
\newblock In \emph{International conference on machine learning}, 2017.

\bibitem[Zadrozny and Elkan(2001)]{zadrozny2001obtaining}
Bianca Zadrozny and Charles Elkan.
\newblock Obtaining calibrated probability estimates from decision trees and naive bayesian classifiers.
\newblock In \emph{Proceedings of the 18th International Conference on Machine Learning}, 2001.

\bibitem[Gneiting et~al.(2007)Gneiting, Balabdaoui, and Raftery]{Gneiting2007}
Tilmann Gneiting, Fadoua Balabdaoui, and Adrian Raftery.
\newblock Probabilistic forecasts, calibration and sharpness.
\newblock \emph{Journal of the Royal Statistical Society: Series B (Statistical Methodology)}, 2007.

\bibitem[Gupta and Ramdas(2021)]{gupta2021distribution}
Chirag Gupta and Aaditya~K. Ramdas.
\newblock Distribution-free calibration guarantees for histogram binning without sample splitting.
\newblock In \emph{Proceedings of the 38th International Conference on Machine Learning}, 2021.

\bibitem[Huang et~al.(2020)Huang, Li, Macheret, Gabriel, and Ohno-Machado]{huang2020tutorial}
Yingxiang Huang, Wentao Li, Fima Macheret, Rodney~A Gabriel, and Lucila Ohno-Machado.
\newblock A tutorial on calibration measurements and calibration models for clinical prediction models.
\newblock \emph{Journal of the American Medical Informatics Association}, 27\penalty0 (4):\penalty0 621--633, 2020.

\bibitem[Wang et~al.(2023)Wang, Joachims, and Gomez-Rodriguez]{wang2022improving}
Lequn Wang, Thorsten Joachims, and Manuel Gomez-Rodriguez.
\newblock Improving screening processes via calibrated subset selection.
\newblock In \emph{Proceedings of the 39th International Conference on Machine Learning}, 2023.

\end{thebibliography}
}

\clearpage
\newpage

\appendix
\section{Additional Experimental Results}
We make two observations matching the theoretical assumptions and results in Corvelo Benz and Gomez Rodriguez~\cite{corvelo2024human} in the study.

Figure~\ref{fig:prob-map} shows the probabilities of guessing red given the initial and final guess of participants. We observe that participants' initial guess is monotonic in their initial confidence, and that the final guess is monotonic in their initial confidence and the AI confidence. 

Given the record of initial confidence $h_{i,j}$, the initial $d_{i,j}$ and final $d'_{i,j}$ color guess made by the participant $j$ in the game instance $i$, we compute the best monotone $\pi_{\text{mono}}$ and best joint $\pi_{\text{joint}}$ guess for all games played. 
The best joint guess maximizes the expected accuracy of the guess given only confidence $h_i$ and $a_i$---the guess is red (black) if the mean probability of a red card for game instances with confidence $h_i$ and $a_i$ is higher (lower) than $50\%$. The best monotone guess maximizes the expected accuracy among guesses made by rational decision makers---guesses that are monotone in confidence $h_i$ and $a_i$.

\begin{figure}
\centering
\subfloat[Expected accuracy averaged across participants and games for each group and type of guess. Red dashed line indicates the expected accuracy of the optimal guess $\pi^*$ and purple dotted line indicates the expected accuracy of the AI playing alone. Both are by design of the study equal across groups and, thus, just drawn as lines.]{
\includegraphics[width=0.8\textwidth]{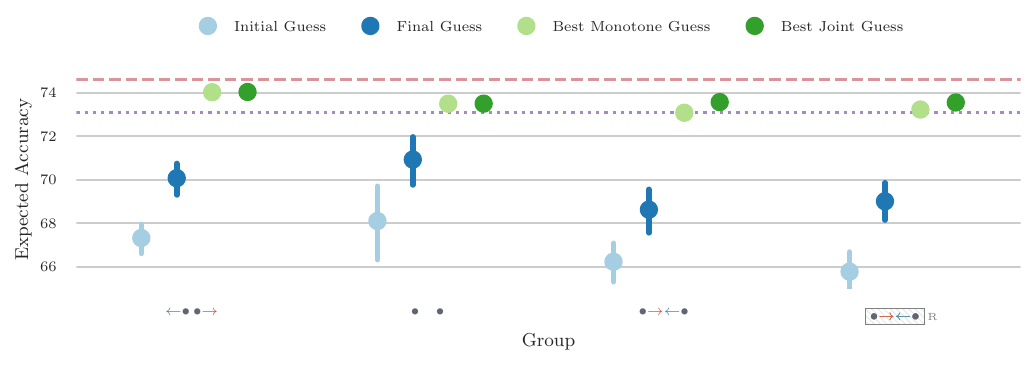}
}
\\
\subfloat[Expected accuracy of each individual participant averaged across games for each group for the initial and final guess.]{
\includegraphics[width=0.8\textwidth]{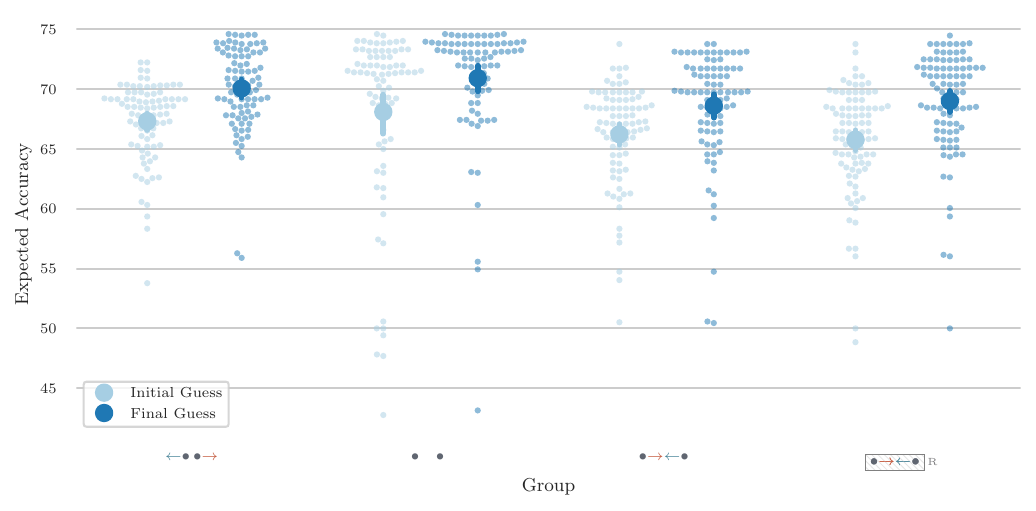}
}
\\
\subfloat[Expected accuracy of each individual participant averaged across games for each group for the best joint and best monotone guess. Some points were not drawn due to space constraints.]{
\includegraphics[width=0.8\textwidth]{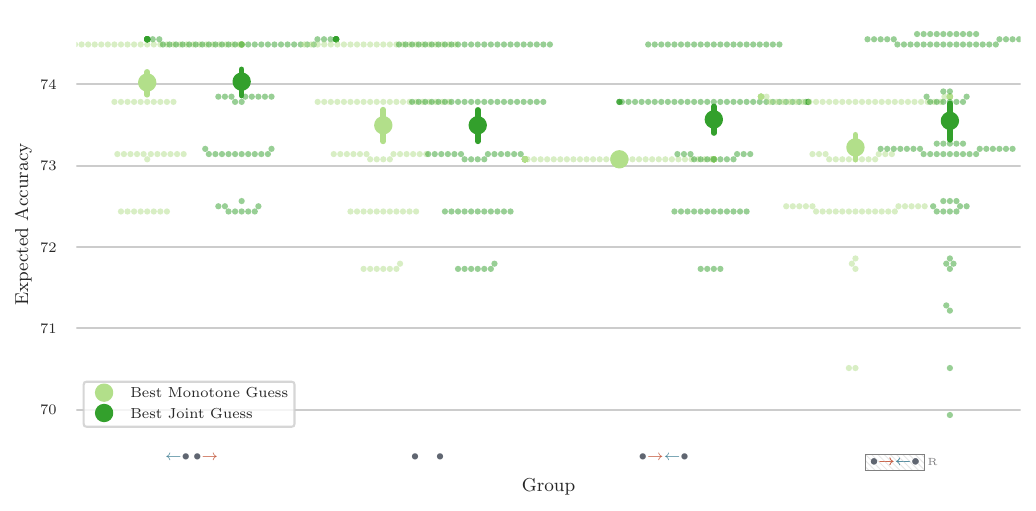}
}
\caption{Accuracy across groups and guesses in expectation. Plot (a) compares means and $95\%$ confidence intervals, while plot (b) and (c) show individual expected accuracies of each participant as well.}
\label{fig:accuracy-app}
\end{figure}

We measure the expected accuracy of each guess averaged across participants and games,
    \begin{equation}\label{eq:exp_acc}
        \sum_{j} \frac{1}{\#\text{participants}} \cdot \sum_{i} \frac{1}{24} \cdot \mathbb{E}_{C_i \sim Be(r_i) }[  \mathds{1}[\pi(i,j) = C_i]]
    \end{equation} 
where $\pi(i,j) \in \{d_{i,j}, d'_{i,j}, \pi_{\text{joint}}(i,j), \pi_{\text{mono}}(i,j)\}$ denotes the guess evaluated.
Figure~\ref{fig:accuracy-app} shows the utility in terms of expected accuracy (Eq.~\ref{eq:exp_acc}) of the best joint guess and the best monotone guess that could be made by the participants based on their own and the AI confidence values.
Although the expected accuracy of the best joint guess is mostly equivalent across groups and the expected accuracy of the AI alone is the same by design, the expected accuracy of the best monotone guess is lower in groups with lower degree of alignment.
However, while the difference in the best monotone guess is barely noteable, the difference in the expected accuracy of the final guess is more prominent. 

\subsection{Expected Calibration Error}
\begin{table}[h!]
    \centering
    \caption{Expected Calibration Error (ECE) measured empirically for each group. As expected, all values are relatively low. The AI in group \protect\symbolB, while empirically the least aligned, also obtains lower ECE than in groups \protect\symbolA and \protect\symbolC. The post-processing algorithm does not significantly reduce ECE of the AI in group \protect\symbolBP.}
    \begin{NiceTabular}{cl}
    \CodeBefore
    \Body
           \bf \small Group & \bf \small ECE  \\
     \hline
     \symbolA & 0.090\\
     \symbolC & 0.055\\
     \symbolB  & 0.036\\
      \ \,\,\symbolBP & 0.033\\
    \hline
    \end{NiceTabular} 
    \label{tab:calibration-error}
\end{table}

\begin{figure*}[t!]
    \centering
    \includegraphics[width=17.8cm]{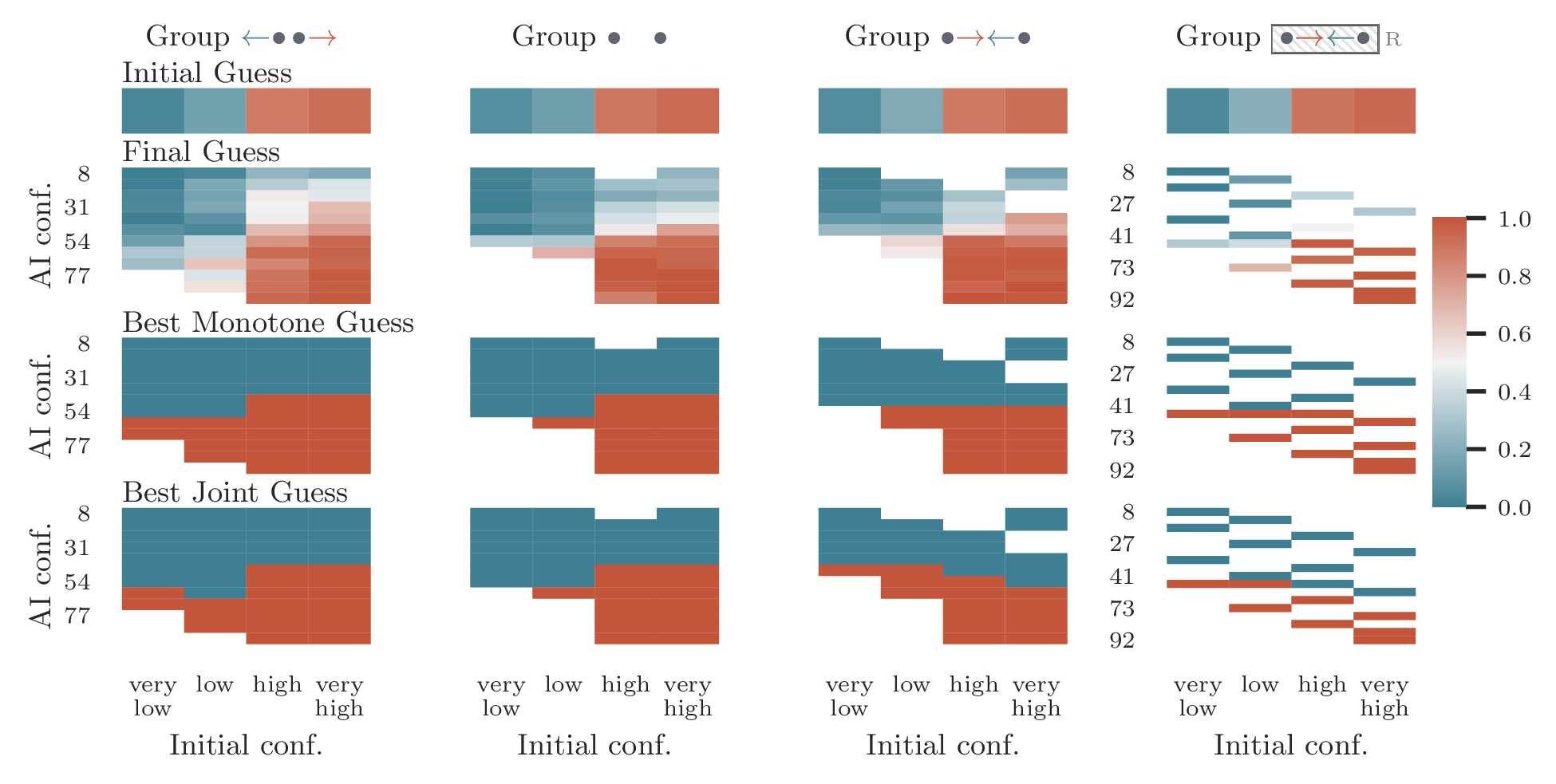}
    \caption{Heatmap of decision probabilities of guessing red stratified by participants' initial confidence and AI confidence shown, and averaged over participants. The initial confidence recorded is discretized into four bins---very low, low, high, very high---denoting the confidence of the participants that the color of the picked card will be red. Bins with 10 or less data points are not displayed. }
    \label{fig:prob-map-app}
\end{figure*}

\begin{figure}
\centering
\subfloat[Heatmap of decision probabilities of guessing red stratified by participants' initial confidence and AI confidence shown, and averaged over participants. The initial confidence recorded is discretized into four bins---very low, low, high, very high---denoting the confidence of the participants that the color of the picked card will be red. Bins with 10 or less data points are not displayed.]{
\includegraphics[width=\textwidth]{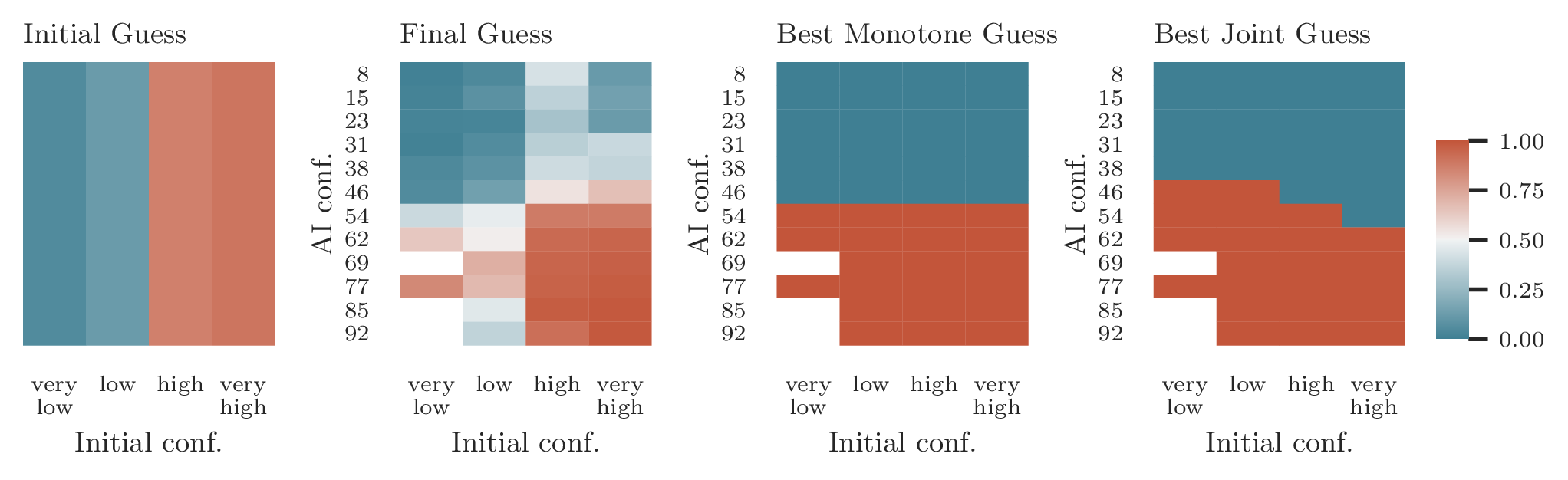}
}
\\
\subfloat[Expected accuracy of each individual participant averaged across games for each guess type. Means over participants are plotted as large circle.]{
\includegraphics[width=\textwidth]{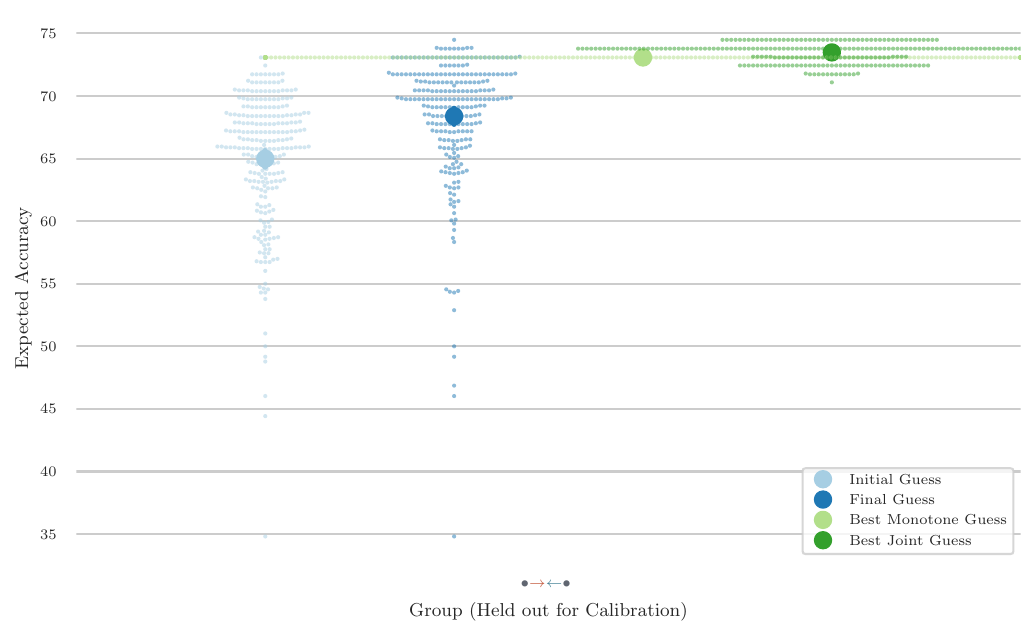}
}
\caption{Summarization plots for held-out calibration data from group \protect \symbolB. Plot (a) shows a heatmap of the probability of guessing red. Plot (b) shows a swarm plot of the expected accuracy of each individual participant averaged across games for each guess type.}
\label{fig:calibration-data}
\end{figure}

\begin{figure}
\centering
\includegraphics[width=\textwidth]{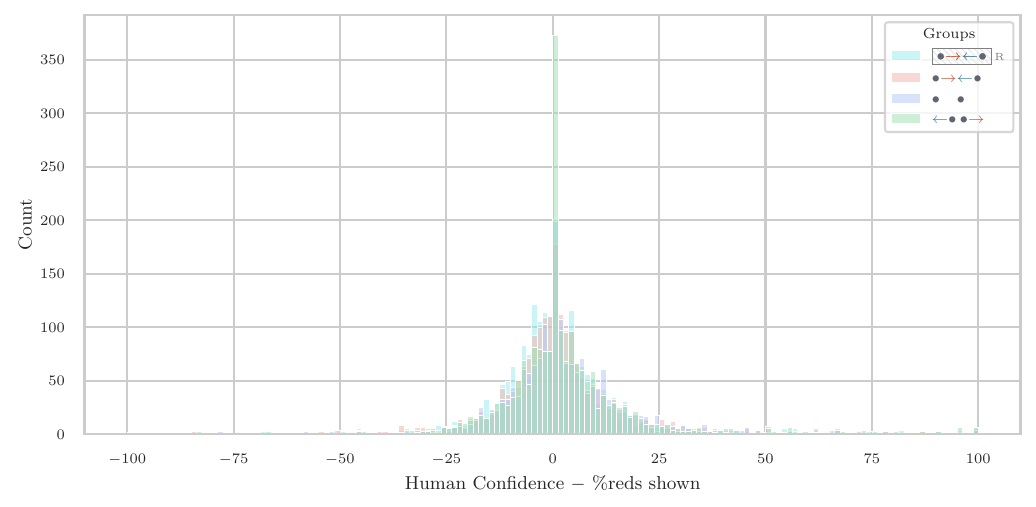}
\caption{Histogram of the difference between reported initial confidence and percentage of reds shown to participants for each group. We observe no difference in distribution across groups. }
\label{fig:histogram}
\end{figure}

\begin{figure}
    \centering
    \includegraphics[width=1\linewidth]{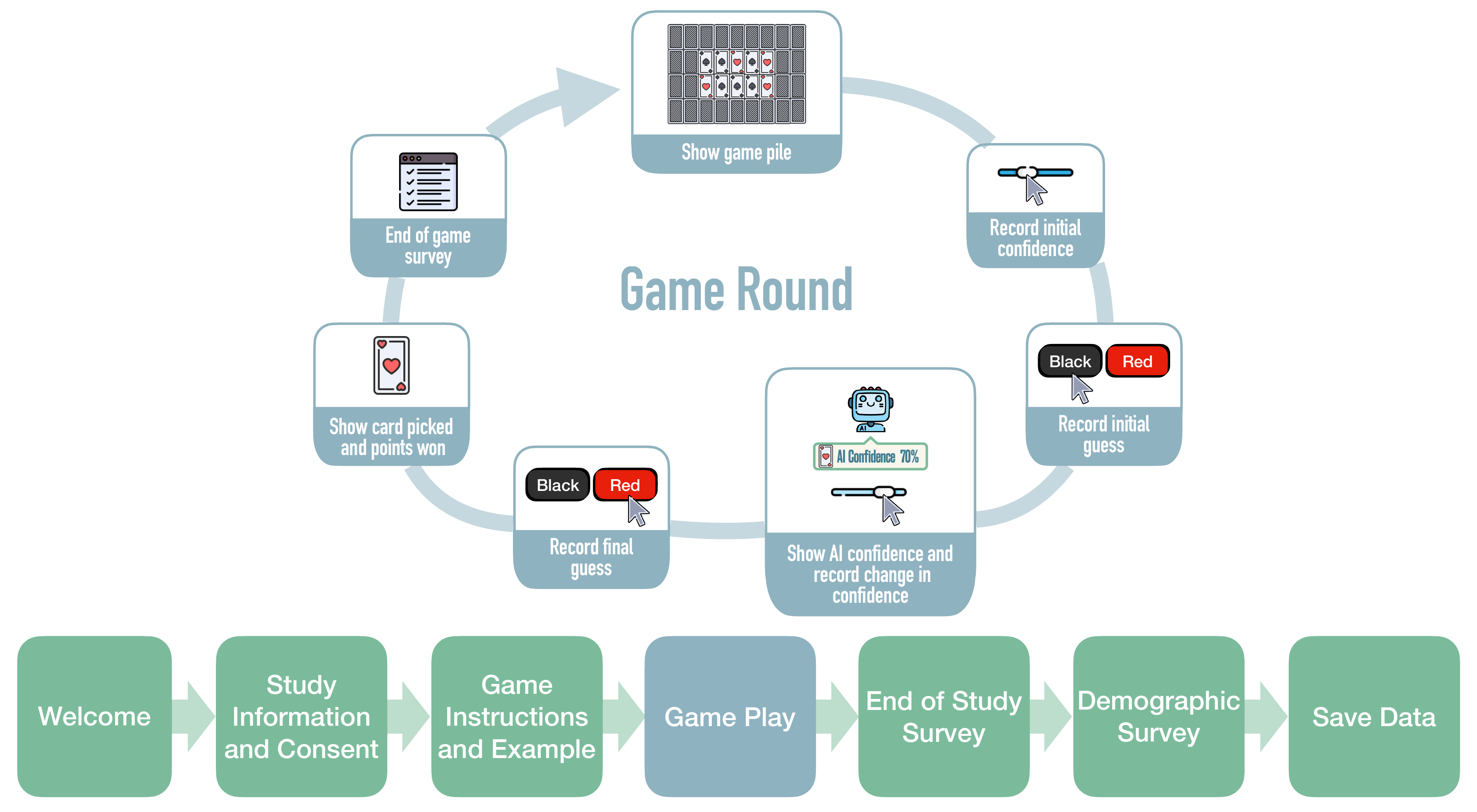}
    \caption{Flow chart showing an overview of the study timeline. Screenshots of an example game round can be found in Figures~\ref{fig:game-pile} to~\ref{fig:outcome}. A screenshot of the end of game survey and end of study survey can be seen in Figure~\ref{fig:surveys}. The full study setup can be found at \url{https://hac-experiment.mpi-sws.org/?PROLIFIC_PID=test&STUDY_ID=test&SESSION_ID=test&LEVEL=B&GAME_BATCH=0}.}
    \label{fig:flowchart}
\end{figure}

\begin{figure}
    \centering
    \includegraphics[width=1\linewidth]{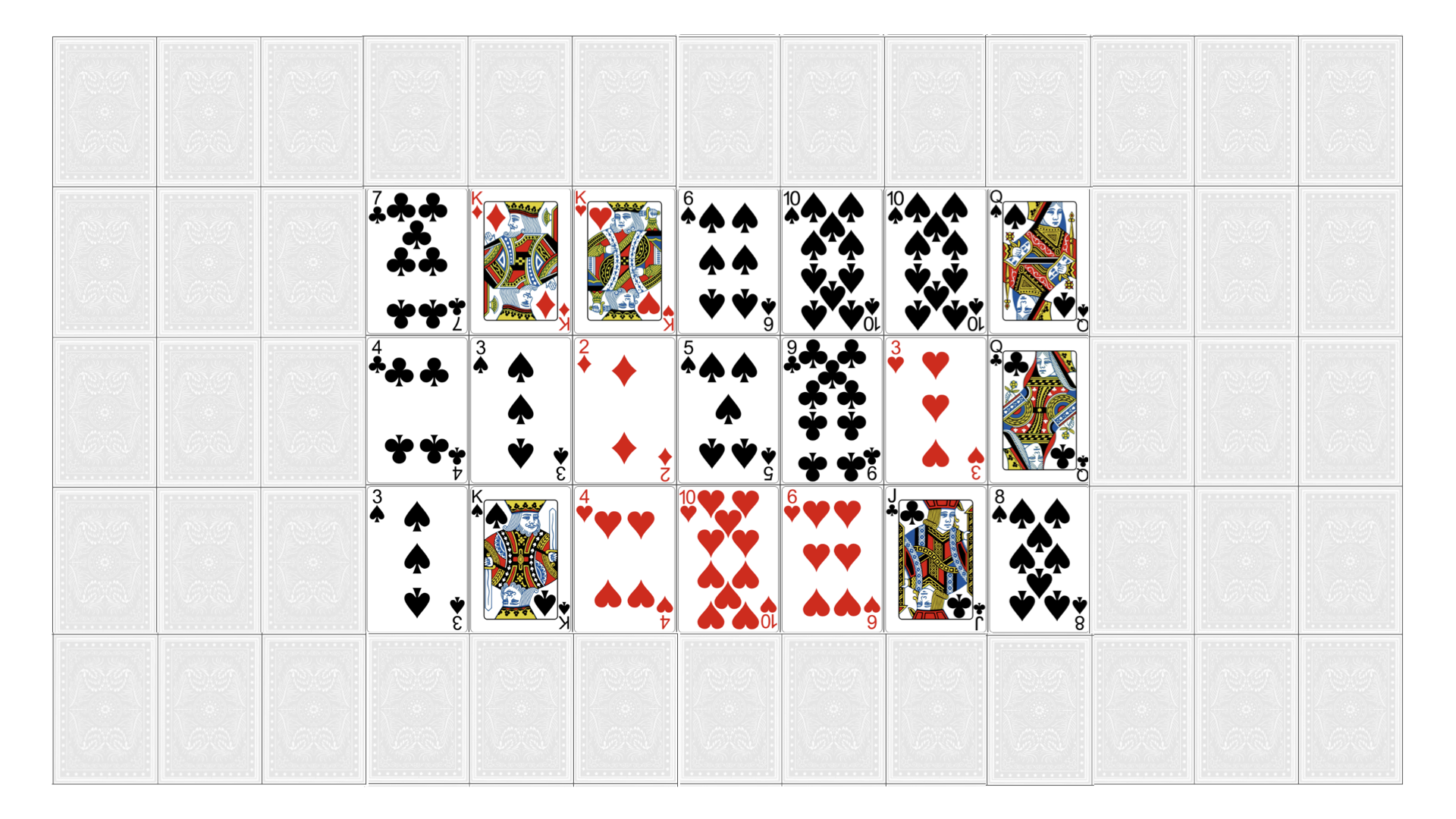}
    \caption{Example of a game pile shown to participants.}
    \label{fig:game-pile}
\end{figure}

\begin{figure}
    \centering
    \subfloat[Record initial confidence]{
    \includegraphics[width=1\linewidth]{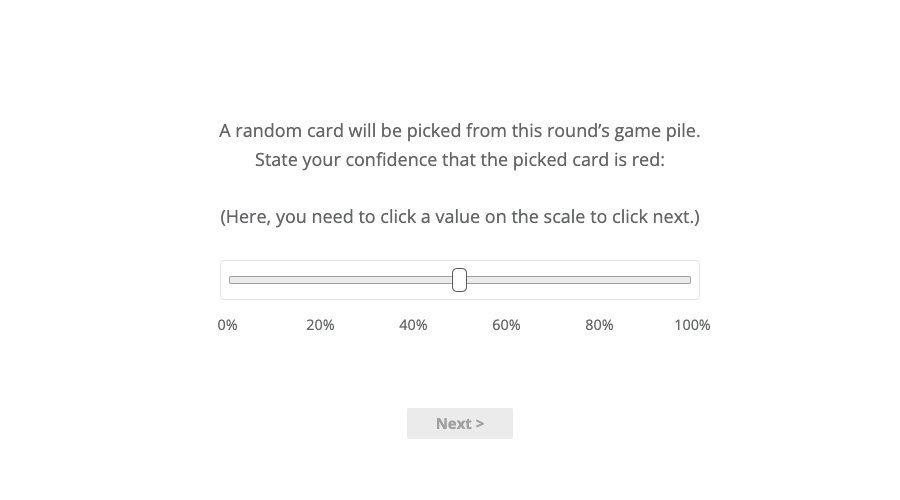}
    }\\
    \subfloat[Record initial guess]{
    \includegraphics[width=1\linewidth]{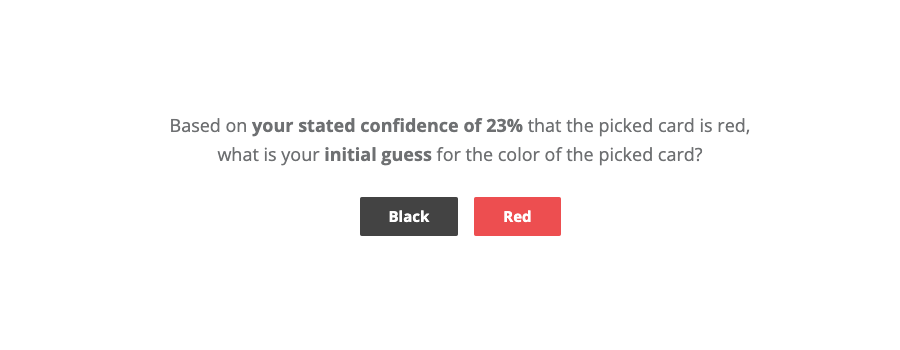}
    }
    \caption{Screenshots of game round steps where participants are asked for their initial confidence and guess before being shown the AI confidence.}
    \label{fig:human-conf}
\end{figure}

\begin{figure}
    \centering
    \subfloat[Show AI confidence and record change in confidence of the participant]{
    \includegraphics[width=1\linewidth]{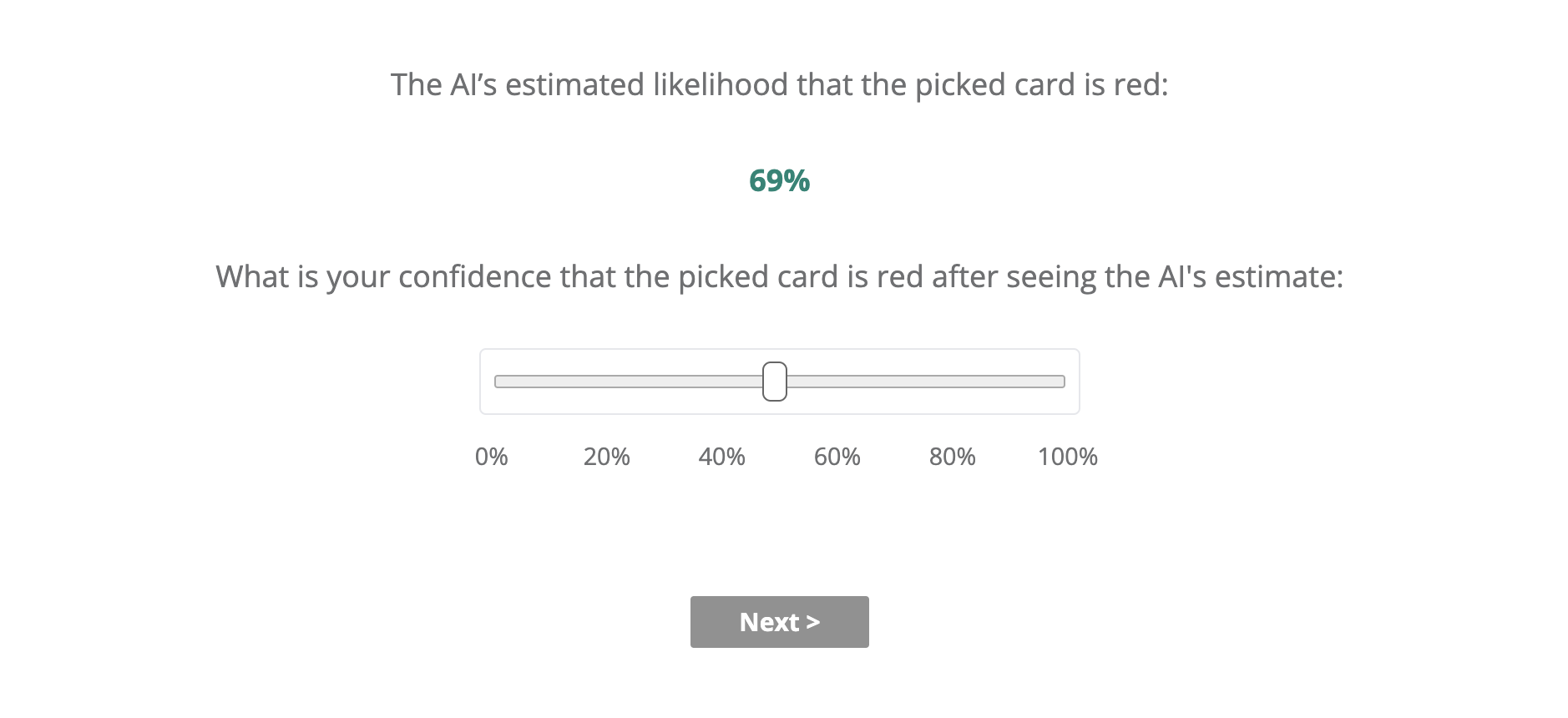}
    }\\
    \subfloat[Record final guess]{
    \includegraphics[width=1\linewidth]{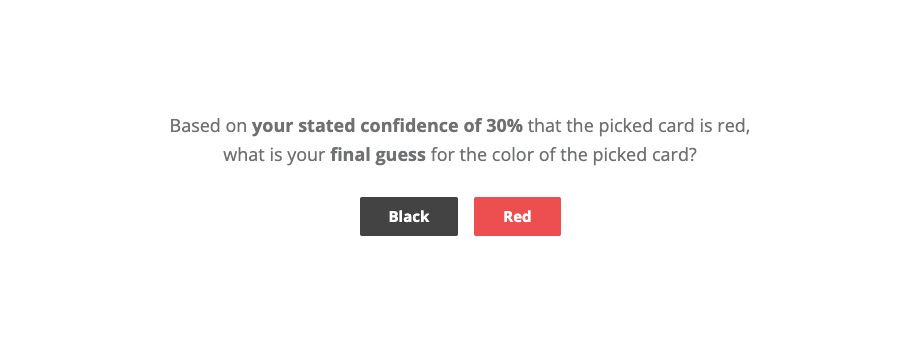}
    }
    \caption{Screenshots of game round steps where participants are shown the AI confidence and asked for their confidence and final guess.}
    \label{fig:AI-conf}
\end{figure}

\begin{figure}
    \centering
    \includegraphics[width=0.75\linewidth]{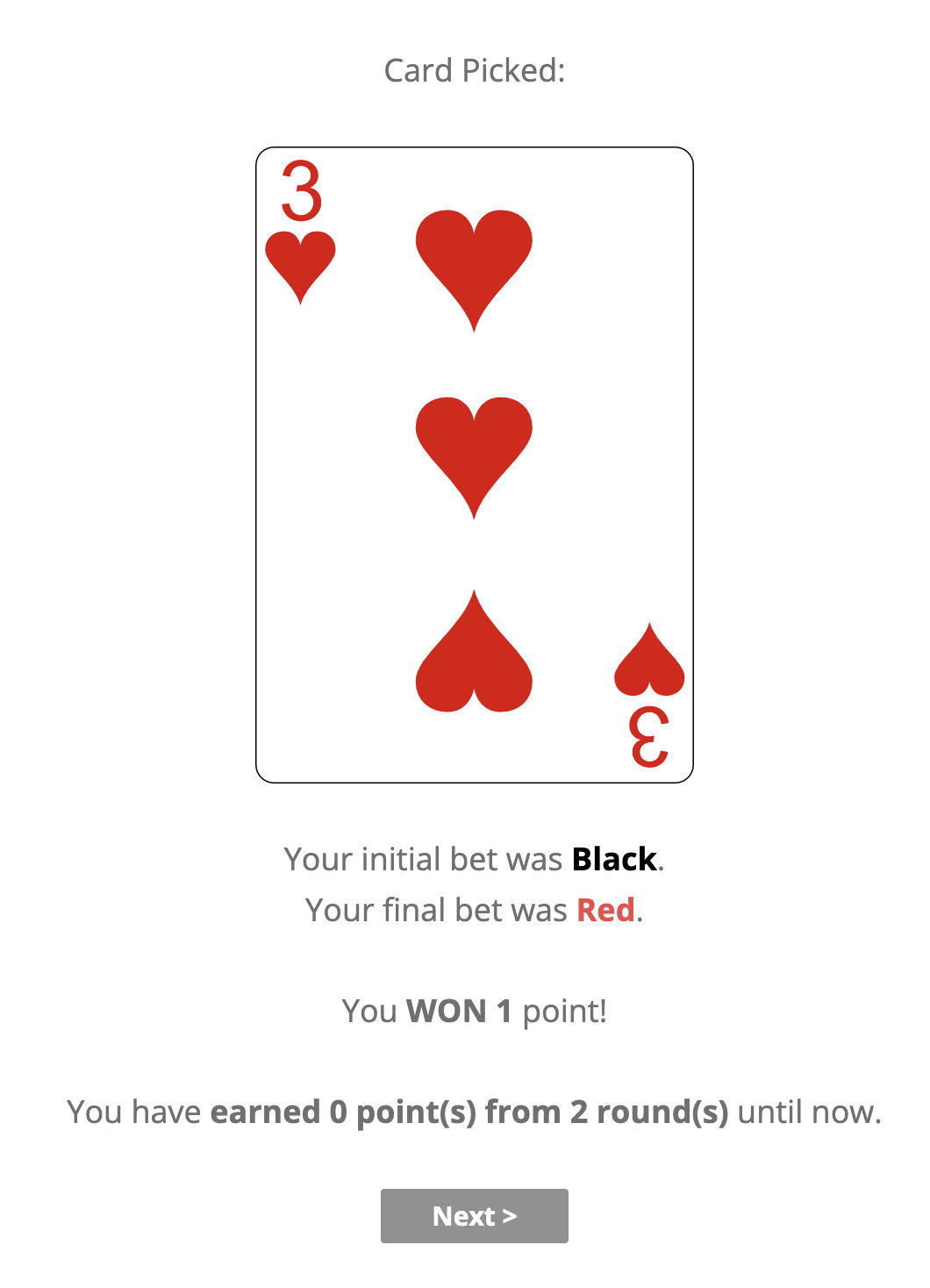}
    \caption{Screenshot of how participants are shown the card picked and points won. }
    \label{fig:outcome}
\end{figure}

\begin{figure}
        \centering
\subfloat[End of game survey]{
        \includegraphics[width=1\linewidth]{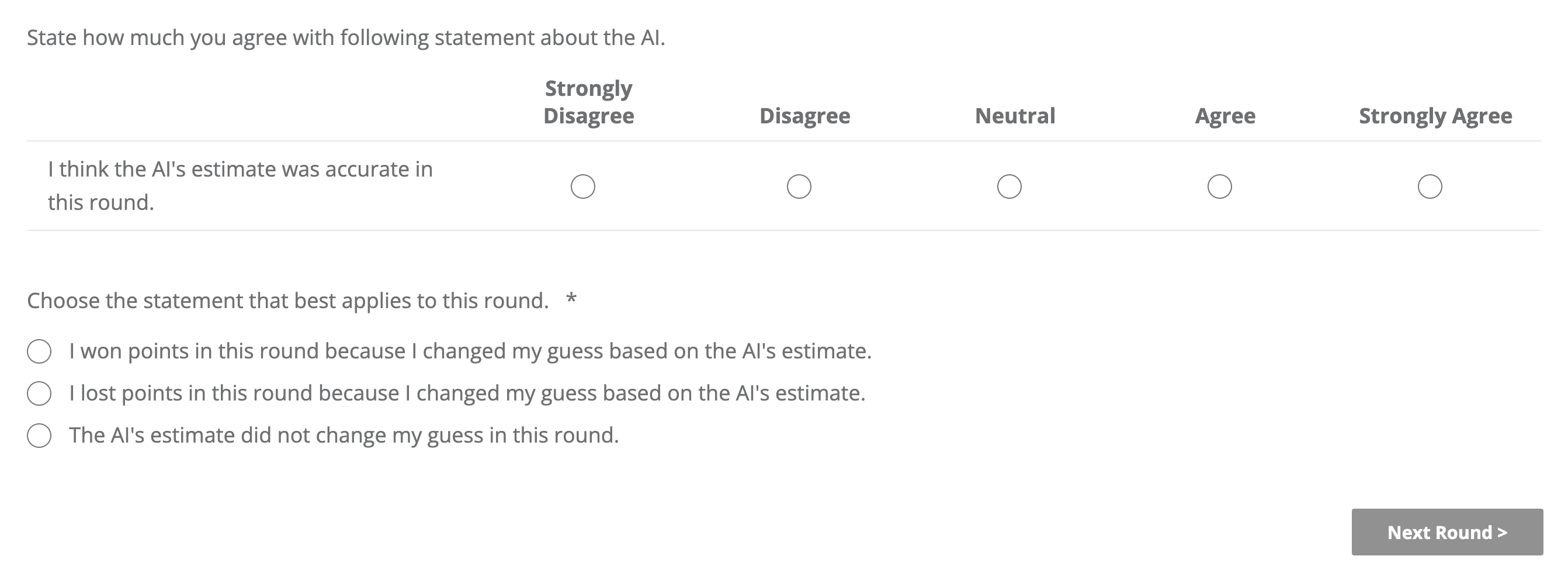}
}
\\
\vspace{2em}
\subfloat[End of study survey]{
    \includegraphics[width=1\linewidth]{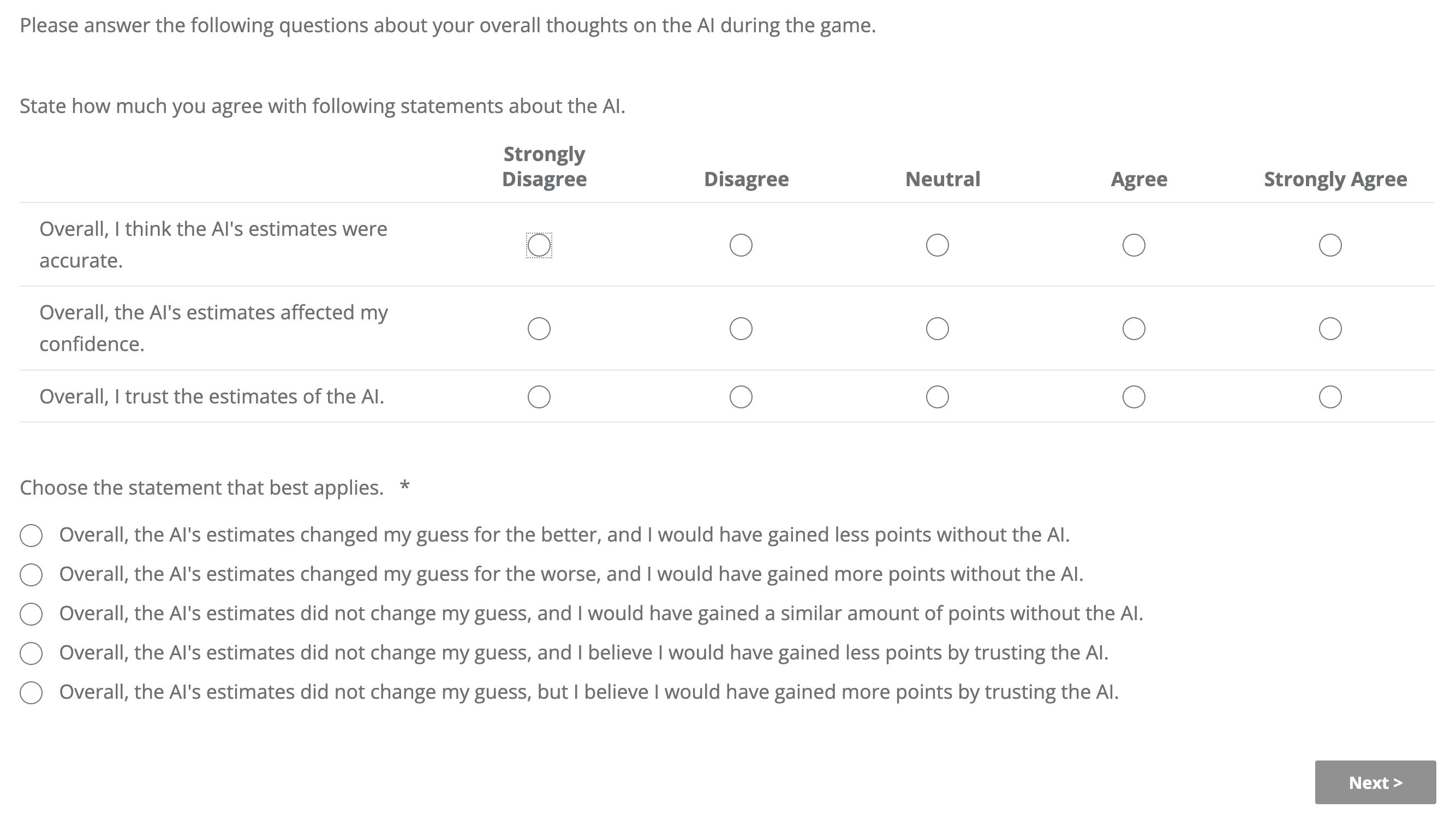}
}
\caption{Screenshot of the end of game and end of study surveys filled out by participants.}
    \label{fig:surveys}
\end{figure}

\begin{figure}
\centering
\subfloat[Averaged value of the likert scale---strongly disagree (0) to strongly agree (5)---in the end of game and end of study survey for following statements:
"I think the AI's estimate was accurate in this round." (Accuracy), "Overall, I think the AI's estimates were accurate." (Overall Accuracy), "Overall, the AI's estimates affected my confidence." (Affected Confidence) and "Overall, I trust the estimates of the AI." (Overall Trust).]{
\includegraphics[width=0.8\textwidth]{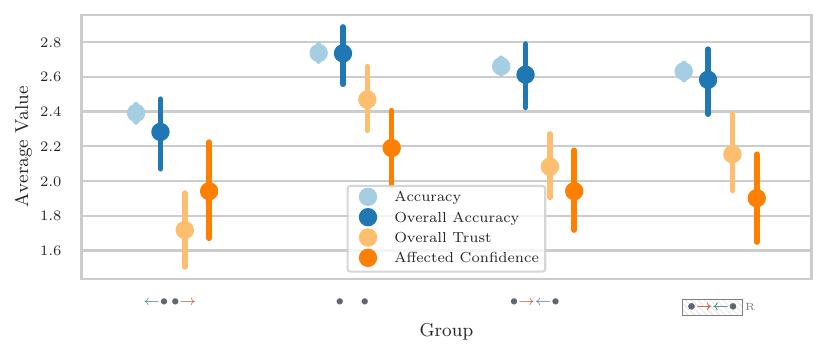}
}
\\
\subfloat[Frequency of statement selected by participants to best apply to their whole game experience at the end of study survey.]{
\includegraphics[width=\textwidth]{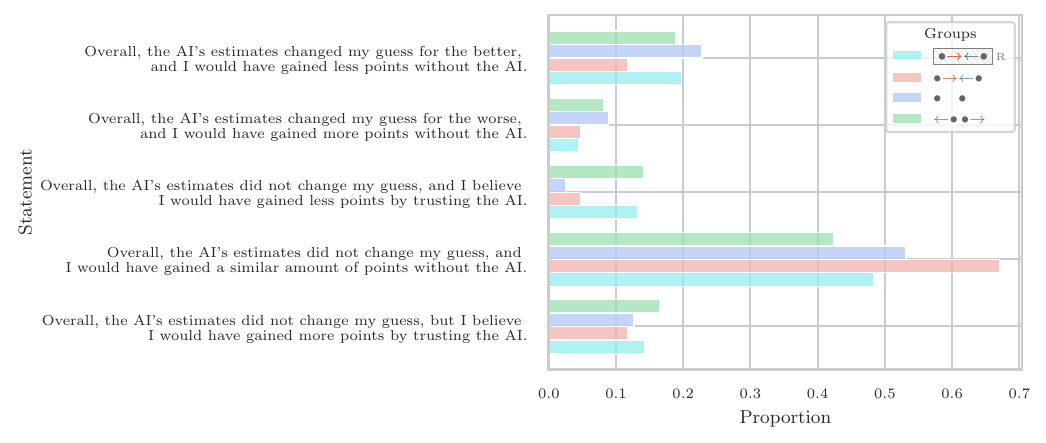}
}
\\
\subfloat[Frequency of statement selected by participants to best apply to their game experience at the end of game survey.]{
\includegraphics[width=\textwidth]{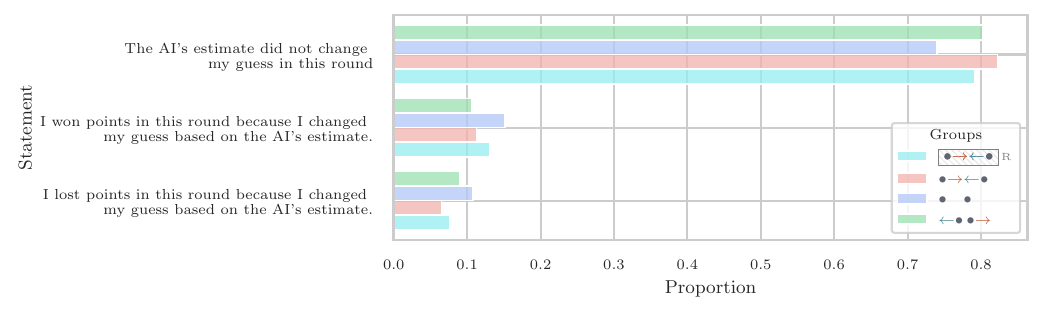}
}
\caption{Summarized responses to end of game and end of study surveys.}
\label{fig:survey_results}
\end{figure}

\begin{figure}
\centering
\includegraphics[width=\textwidth]{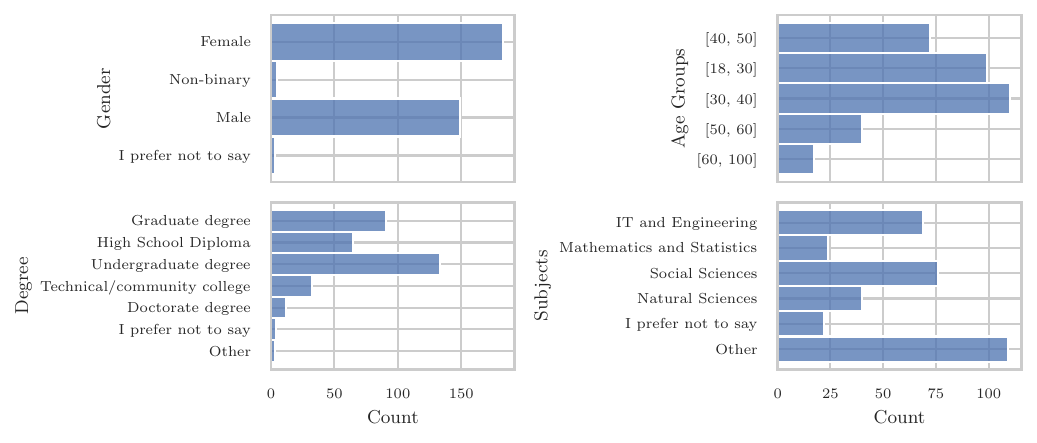}
\caption{Demographic information provided by participants including gender, age, highest obtained degree and subject of degree.}
\label{fig:demographic}
\end{figure}

\end{document}